\documentclass[11pt,a4paper]{article}
\usepackage[hyperref]{eacl2021}
\usepackage{times}
\usepackage{url}

\usepackage{latexsym}

\usepackage{microtype}

\usepackage{times}
\usepackage{latexsym}
\usepackage{lingmacros}
\usepackage{amsmath}
\usepackage{graphicx}
\usepackage{array, tabularx, caption, boldline}
\usepackage{cellspace}
\usepackage{url}
\usepackage{makecell}
\usepackage{multirow}
\usepackage{xcolor}

\usepackage{subcaption}
\usepackage{booktabs}
\usepackage{comment}

\usepackage{pifont}
\newcommand{\cmark}{\ding{51}}%
\newcommand{\xmark}{\ding{55}}%

\aclfinalcopy 

\title{Two Training Strategies for Improving Relation Extraction over Universal Graph}

\author{Qin Dai$^1$, Naoya Inoue$^{2}$, Ryo Takahashi$^{1,3}$, Kentaro Inui$^{1,3}$ \\ $^1$Tohoku University, Japan \\ $^2$Stony Brook University, USA \\ $^3$RIKEN Center for Advanced Intelligence Project, Japan \\ {\tt \{daiqin, naoya-i, ryo.t, inui\}@ecei.tohoku.ac.jp}}

\date{}

\begin{document}
\maketitle
\begin{abstract}
   This paper explores how the Distantly Supervised Relation Extraction (DS-RE) can benefit from the use of a Universal Graph (UG), the combination of a Knowledge Graph (KG) and a large-scale text collection. A straightforward extension of a current state-of-the-art neural model for DS-RE with a UG may lead to degradation in performance. We first report that this degradation is associated with the difficulty in learning a UG and then propose two training strategies: (1) Path Type Adaptive Pretraining, which sequentially trains the model with different types of UG paths so as to prevent the reliance on a single type of UG path; and (2) Complexity Ranking Guided Attention mechanism, which restricts the attention span according to the complexity of a UG path so as to force the model to extract features not only from simple UG paths but also from complex ones. Experimental results on both biomedical and NYT10 datasets prove the robustness of our methods and achieve a new state-of-the-art result on the NYT10 dataset. The code and datasets used in this paper are available at \url{https://github.com/baodaiqin/UGDSRE}.

\end{abstract}

\section{Introduction}
\label{sec:intro}
Relation Extraction (RE) is an important task in Natural Language Processing (NLP). RE aims to turn unstructured texts into structured Knowledge Graph (KG), which is typically stored as ($e_1$, $r$, $e_2$) triplets, where $e_1$ is a \textit{head entity}, $r$ is a relation and $e_2$ is a \textit{tail entity}, such as (\textit{aspirin}, \textit{may\_treat}, \textit{pain}) and (\textit{Guy Maddin}, \textit{place\_lived}, \textit{Winnpeg}). RE can be formulated as a classification task to predict a predefined relation $r$ from entity pair $(e_1,e_2)$ annotated evidences.

One obstacle that is encountered when building a RE system is the generation of a large amount of manually annotated training instances, which is expensive and time-consuming. For coping with this difficulty, \newcite{mintz2009distant} propose Distant Supervision (DS) to automatically generate training samples via linking KGs to texts. They assume that if ($e_1$, $r$, $e_2$) is in a KG, then all sentences that contain ($e_1$, $e_2$) (hereafter, \emph{sentence evidences}) express the relation $r$. It is well known that the DS assumption is too strong and inevitably accompanies the wrong labeling problem, such as the sentence evidences (\ref{sent:wronglabel} and \ref{sent:wronglabel_gen}) below, which fail to express \textit{may\_treat} and \textit{place\_lived} relation respectively.
\eenumsentence{
    \item {\it \textbf{Aspirin}$_{e_1}$ is widely used for short-term treatment of \textbf{pain}$_{e_2}$, fever or colds.} \label{sent:goodlabel}
    \item {\it The tumor was remarkably large in size, and \textbf{pain}$_{e_2}$ unrelieved by \textbf{aspirin}$_{e_1}$.} \label{sent:wronglabel}
}
\enumsentence{
    {\it He is now finishing a documentary about \textbf{Winnipeg}$_{e_2}$, the final installment of a personal trilogy that began with “Cowards Bend the Knee” (a 2003 film that also featured a hapless hero named \textbf{Guy Maddin}$_{e_1}$).}
    \label{sent:wronglabel_gen}
}

Recently, neural network models with attention mechanism have been proposed to alleviate the wrong labeling problem and attend informative sentence evidences such as (\ref{sent:goodlabel})~\cite{lin2016neural,ji2017distant,du2018multi,jat2018improving,han2018neural,han2018hierarchical}.
However, there can be a large portion of entity pairs that lack such informative sentence evidences that explicitly express their relation.
This makes Distantly Supervised Relation Extraction (DS-RE) further challenging~\cite{sun2019leveraging}.
\begin{figure*}[t]
\centering
\includegraphics[width=\columnwidth*2]{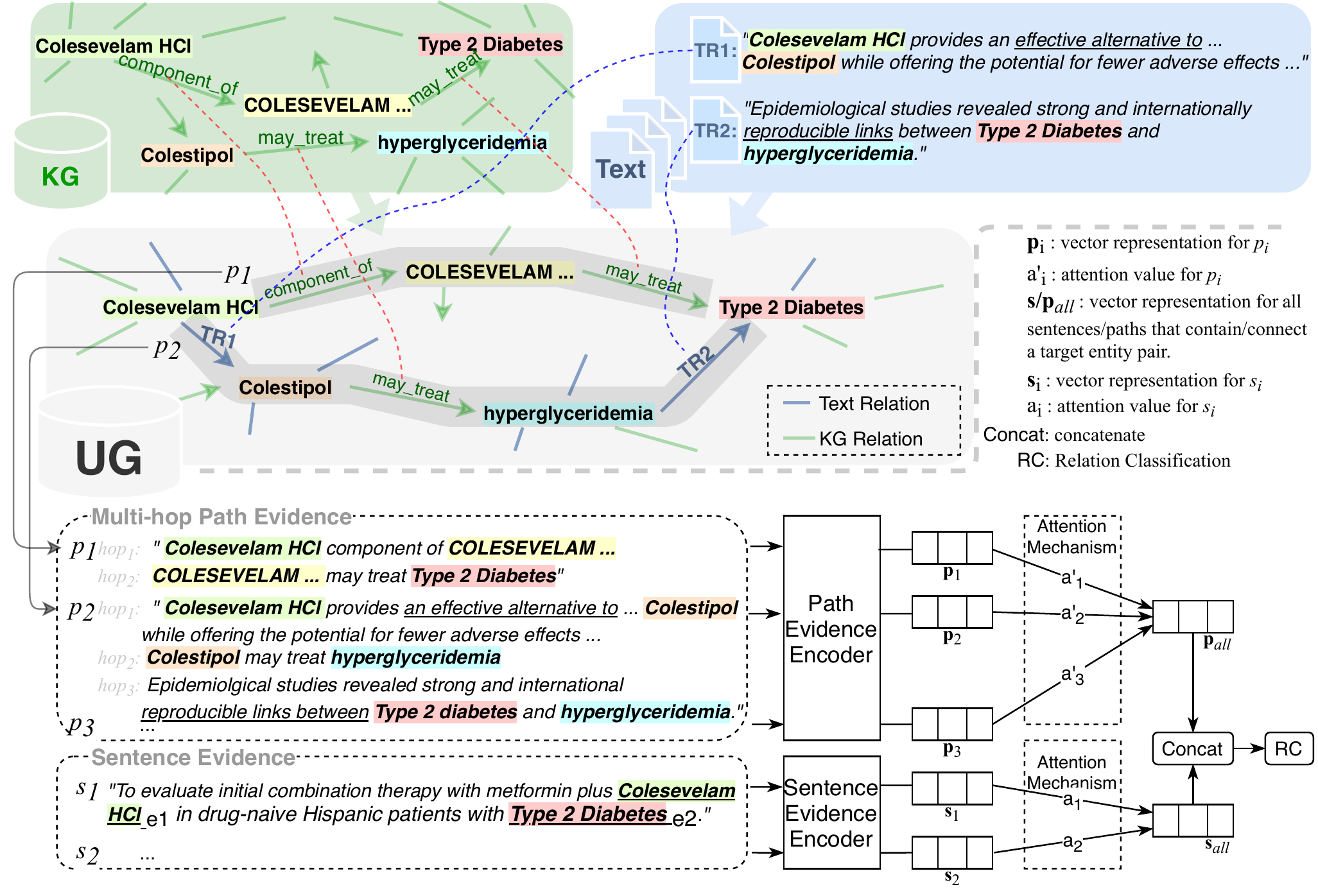}
\caption{Overview of our UG-based framework, where \textbf{Colesevelam HCl} and \textbf{Type 2 Diabetes} are the target entities, \textbf{COLESEVELAM ...}, \textbf{Colestipol} and \textbf{hyperglyceridemia} are intermediate entities, each UG path consists of multiple hops and each hop represents a KG relation (such as ``\textit{Colestipol may treat hyperglyceridemia}'') or Text (or Textual) relation (such as \textsf{TR1} and \textsf{TR2}), which is the sentence containing two (target or intermediate) entities.}
\label{fig:ugexam}
\end{figure*}

For compensating the lack of informative sentence evidences, \newcite{dai2019incorporating} utilize multi-hop paths connecting a target entity pair (hereafter, \emph{path}) over a KG as extra evidences for DS-RE.
An example of such multi-hop KG path can be seen in Figure~\ref{fig:ugexam}, where $p_1$ depicts a multi-hop KG path of the form of {\it $e_1$ $\underrightarrow{\textit{component\_of}}$  $e_3$ $\underrightarrow{\textit{may\_treat}}$ $e_2$}.
The model of \newcite{dai2019incorporating} uses such multi-hop paths as additional features for predicting the relation between a given target entity pair ($e_1$, $e_2$), which is reported effective for performance improvement.
However, KGs are often highly incomplete~\cite{min2013distant} and may be too sparse to provide enough informative paths in practice, which may hamper the effectiveness of multi-hop paths. 

Given this background, in this study, we take one step further, aiming for inducing maximal signals of distant supervision from both a KG and a large text collection (hereafter, \emph{Text}).
For this purpose, we consider using multi-hop paths over a Universal Graph (UG) as extra features for DS-RE. 
Here, we define a UG as a joint graph representation of both KG and Text, where each node represents an entity from KG or Text, and each edge indicates a KG relation or Textual relation, as shown in Figure~\ref{fig:ugexam}. 
The path $p_2$ in the figure is an example of UG path, comprising a textual edge \textsf{TR1}, a KG edge \textit{may\_treat}, and another textual edge \textsf{TR2}. 
By augmenting the original KG with textual edges, one can expect far more chances to find informative path evidences between any given target entity pairs, because the number of such textual edges is likely to be much larger than the number of KG edges (Note that one can collect as many textual edges as needed from a raw text corpus with an entity linker).
Extending a KG to a UG, therefore, may allow a DS-RE model to learn richer distant supervision signals. 

The idea of using multi-hop paths over a UG is not necessarily new on its own. 
For example, \newcite{toutanova2015representing} propose to use a UG for knowledge graph completion, and \newcite{das2017question} propose a model trained to reason over a UG for question answering. 
However, there is no prior study that has explored the effective way to use a UG for the task of DS-RE from text. 
In fact, finding an effective way of using a UG for DS-RE is not as simple as it may seem. 
As we report in this paper, a straightforward extension of the \newcite{dai2019incorporating} model to the UG setting may result in performance degradation. 

Motivated by this, in this paper, we address how one can make effective use of UG for DS-RE. 
We first report our observation that a straightforward extension of the \newcite{dai2019incorporating} model to the UG setting tends to allocate the majority of attention to only a limited set of UG paths such as short KG paths and miss out the learning from a wide range of UG paths (\S\ref{sec:attbia}), which hinders performance gain.
In order to alleviate the negative effect of the attention bias and realize the potential of UG paths, we propose two training  (or debiasing) strategies: (1) Path Type Adaptive Pretraining (\S\ref{sec:pre}), which aims to improve the adaptability of the model to various UG paths; and (2) Complexity Ranking Guided Attention mechanism (\S\ref{sec:rank}), which enables the model to learn from both simple and complex UG paths. Experimental results on both biomedical and NYT10~\cite{riedel2010modeling} datasets prove that: (1) UG paths have the potential to bring performance gain for DS-RE as compared with KG paths; (2) the proposed training methods are effective to fully exploit the potential of UG paths for DS-RE because the proposed methods significantly and consistently outperform several baselines on both datasets and especially achieve a new state-of-the-art result on the NYT10 dataset.

\section{Related Work}

To improve the performance of a DS-RE model, recently, researchers introduce various attention mechanisms. \newcite{lin2016neural} propose a relation vector based attention mechanism. \newcite{jat2018improving,du2018multi} propose multi-level (e.g., word-level and sentence-level) structured attention mechanism. \newcite{ye2019distant} apply both intra-bag and inter-bag attention for DS-RE. \newcite{han2018hierarchical} propose a relation hierarchy based attention mechanism. \newcite{han2018neural} propose a joint model that adopts a KG embeddings based attention mechanism. \newcite{jia2019arnor} propose an attention regularization framework to select informative sentence evidences for DS-RE. However, these models rely only on noisy sentence evidences from DS, neglecting the rich UG paths for DS-RE.

Besides the sentence evidences from DS, researchers also leverage external evidences for DS-RE. 
\newcite{ji2017distant} apply entity descriptions generated from Freebase and Wikipedia as extra evidences, \newcite{lin2017neural} utilize multilingual text as extra evidences and \newcite{vashishth2018reside} use multiple extra evidences including entity types, dependency and relation alias information for DS-RE. \newcite{alt2019fine} utilize pretrained language model as background information for DS-RE. \newcite{sun2019leveraging} apply relational table extracted from Web as supplementary evidences for DS-RE. 

To apply DS-RE beyond sentence boundary, \newcite{quirk2017distant} utilize syntactic information to extract relation from neighboring sentences. \newcite{zeng2017incorporating} apply two-hop KG paths
identified from two-hop textual paths as extra evidences for DS-RE. Different from this work, we directly use the rich UG paths as extra evidences.
\newcite{dai2019incorporating} extend the framework of \newcite{han2018neural} by introducing multiple KG paths as extra evidences for DS-RE. \newcite{neelakantan-etal-2015-compositional,das2017chains} use multiple reasoning paths over Text and KG for relation prediction in the paradigm of Knowledge Graph Completion. Our work differs from the ones mentioned above in two ways: (i) We utilize the UG paths as extra evidences for the task of DS-RE from text, (ii) We take into account the factor of attention bias while encoding UG paths and propose two effective debiasing methods to exploit the potential of UG paths for DS-RE.

\section{Base Model}
\label{sec:evencoder} 
\begin{figure}[t]
\centering
\includegraphics[width=\columnwidth]{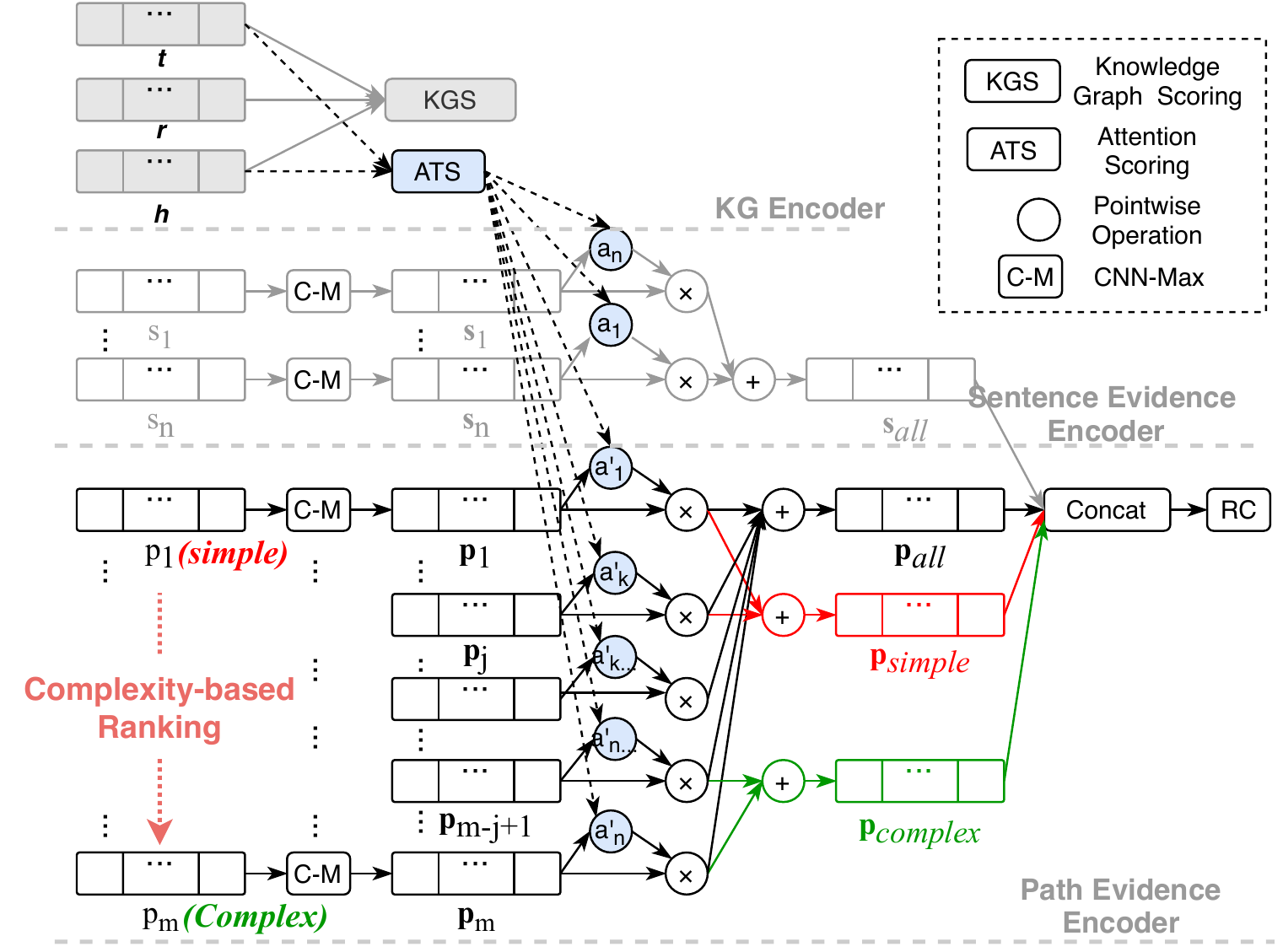}
\caption{Overview of the base model and our proposed Complexity Ranking Guided Attention mechanism (\S\ref{sec:rank}). The base model takes the sentence evidences (e.g., $s_1$, ...) containing a target entity pair and the UG paths (e.g., $p_1$, ...) connecting the entity pair as input for predicting their relation. The KG embeddings of the entity pair (i.e., $\mathbf{h}$ and $\mathbf{t}$) are used for calculating the attention over these sentences and paths. The Complexity Ranking Guided Attention mechanism is proposed to force the model to attend both simple UG paths (e.g., $p_1 \sim p_j$) and complex ones (e.g., $p_{m-j+1} \sim p_m$).}
\label{fig:proposed_model}
\end{figure}

We select the DS-RE model proposed by \newcite{dai2019incorporating} as our base model and extend it into our UG setting. Given a target entity pair $(e_1, e_2)$, a bag of corresponding sentence evidences $S_r=\{s_1,...,s_n\}$ and a bag of UG paths $P_r=\{p_1,...,p_m\}$, the base model aims to measure the probability of $(e_1, e_2)$ having a predefined relation $r$ (including the empty relation NA).
The base model consists of four main modules: KG Encoder, Sentence Evidence Encoder, Path Evidence Encoder and Relation Classification Layer, as shown in Figure~\ref{fig:proposed_model}.

\subsection{KG Encoder} Suppose we have a KG containing a set of fact triplets $\mathcal{O}=\{(e_1, r, e_2), ...\}$, where each fact triplet consists of two entities $e_1, e_2 \in \mathcal{E}$ and their relation $r \in \mathcal{R}$. Here $\mathcal{E}$ and $\mathcal{R}$ stand for the set of entities and relations respectively.

The KG Encoder then encodes $e_1, e_2 \in \mathcal{E}$ and their relation $r\in\mathcal{R}$ into low-dimensional vectors $\mathbf{h}$, $\mathbf{t}$ $\in R^d$ and $\mathbf{r}$ $\in R^d$ respectively, where $d$ is the dimensionality of the embedding space. The KG Encoder adopts TransE~\cite{bordes2013translating} to score a given triplet. Specifically, given a triplet $(e_1, r, e_2)$, TransE evaluates its plausibility via Equation~\ref{eq:kgc2}:
\begin{align}
f_r(e_1,e_2) & = b - \|\mathbf{r}_{ht}-\mathbf{r}\|, \label{eq:kgc2} \\
\mathbf{r}_{ht} & = \mathbf{t} - \mathbf{h}, \label{eq:transe1}
\end{align}
where $b$ is a bias constant and $\mathbf{r}_{ht}$ is a latent relation embedding for $(e_1, e_2)$.
The conditional probability can be formalized over all fact triplets $\mathcal{O}$ as follows:
\begin{equation}
\begin{split}
\mathit{P}(e_1,r,e_2|\theta_{\mathcal{E}}, \theta_{\mathcal{R}}) =
\frac{\exp(f_{r}(e_1,e_2))}{\sum_{r'\in \mathcal{R}}\exp(f_{r'}(e_1,e_2))}
\end{split}
\end{equation}
where $\theta_{\mathcal{E}}$ and $\theta_{\mathcal{R}}$ are parameters for entities and relations respectively. 

\subsection{Sentence Evidence Encoder} Given a bag of sentence evidences $S_r=\{s_1,...,s_n\}$, the Sentence Evidence Encoder applies CNN-Max (see Appendix \S\ref{ap:cnnmax}) on each sentence, namely $\mathbf{s}_i = \text{CNN-Max}(s_i)$, to derive the sentence representations $\{\mathbf{s}_1,...,\mathbf{s}_n\}$.
The encoder then calculates the bag-level vector representation $\mathbf{s}_{\mathrm{all}}$ via Equation~\ref{eq:cnnall}:
\begin{align}\label{eq:cnnall}
\mathbf{s}_{\mathrm{all}} &= \sum^n_{i=1}a_i\mathbf{s}_i,\\ 
a_i &=\frac{\exp(\langle\mathbf{r}_{ht},\mathbf{x}_i\rangle)}
{\sum^n_{k=1}\exp(\langle\mathbf{r}_{ht},\mathbf{x}_k\rangle)},\nonumber\\
\mathbf{x}_i &= \tanh(\mathbf{W}\mathbf{s}_i + \mathbf{b}) \nonumber
\end{align}
where $\mathbf{r}_{ht}$ is from Equation~\ref{eq:transe1}, $\mathbf{W}$ is the weight matrix, $\mathbf{b}$ is the bias vector, $a_i$ is the weight for the $i$-th sentence in $S_r$.

\subsection{Path Evidence Encoder} Given a bag of UG paths $P_r=\{p_1,...,p_m\}$ connecting an entity pair of interest ($e_1$, $e_2$), the Path Evidence Encoder encodes them into a bag-level vector representation $\mathbf{p}_{\mathrm{all}}$. Since we represent a path as a sequence of words (or a long sentence), as shown in Figure~\ref{fig:ugexam}, analogously to the Sentence Evidence Encoder, we apply a CNN-Max (see Appendix \S\ref{ap:cnnmax}) to encode each path $p_i$, namely $\mathbf{p}_i = \text{CNN-Max}(p_i)$. The bag-level path representation $\mathbf{p}_{\mathrm{all}}$ for $P_r$ is then calculated via Equation~\ref{eq:pathall}:
\begin{align}\label{eq:pathall}
\mathbf{p}_{\mathrm{all}} &=\sum^m_{i=1}a'_i\mathbf{p}_i,\\
a'_i &= \frac{\exp(\langle\mathbf{r}_{ht},\mathbf{x'}_i\rangle)}
{\sum^m_{k=1}\exp(\langle\mathbf{r}_{ht},\mathbf{x'}_k\rangle)},\nonumber\\
\mathbf{x'}_i &= \tanh(\mathbf{W}\mathbf{p}_i + \mathbf{b}) \nonumber
\end{align}
where $a'_i$ is the weight for the $i$-th path in $P_r$.

\subsection{Relation Classification Layer} The conditional probability of $(e_1,e_2)$ having a relation $r$ is formulated via Equation~\ref{eq:pathall11}:
\begin{equation}\label{eq:pathall11}
P(e_1,r,e_2|S_r,P_r,\theta_S, \theta_P)=\frac{\exp([\mathbf{o}]_r)}
{\sum_{c=1}^{n_r}\exp([\mathbf{o}]_c)}
\end{equation}
%
%
%
where $\mathbf{o}=\mathbf{M}[\mathbf{s}_{\mathrm{all}}; \mathbf{p}_{\mathrm{all}}]+\mathbf{d}$, $\theta_S$, $\theta_P$ are the parameters in Sentence Evidence Encoder and Path Evidence Encoder, $\mathbf{M}$ is the representation matrix of relations, $\mathbf{d}$ is a bias vector, $\mathbf{o}$ is the output vector containing the prediction scores of all predefined relations, $[\mathbf{o}]_c$ is the prediction score for the relation $c$, and $n_r$ is the total number of relations.

Given a training dataset consisting of triplets $\mathcal{O}=\{(e_1^1, r^1, e_2^1), (e_1^2, r^2, e_2^2), ...\}$, we minimize the objective function as follows:
\begin{equation}\label{eq:objective_prop}
\begin{aligned}
J(\theta)=-\frac{1}{|\mathcal{O}|}\sum_{i=1}^{|\mathcal{O}|}\log P(e_1^i,r^i,e_2^i|\theta_{\mathcal{E}},\theta_{\mathcal{R}}) \\ + \log P(e_1^i,r^i,e_2^i|S_{r^i}, P_{r^i}, \theta_S,\theta_P)
\end{aligned}
\end{equation}
The base model is optimized with Stochastic Gradient Descent (SGD). Following \cite{han2018neural}, we optimize $P(e_1^i,r^i,e_2^i|\theta_{\mathcal{E}},\theta_{\mathcal{R}})$ and $P(e_1^i,r^i,e_2^i|S_i, P_i, \theta_S,\theta_P)$ in parallel.

\section{Proposed Training Method}
\label{sec:proposed}

\subsection{Problem of Attention Bias}
\label{sec:attbia}

\begin{figure*}
\centering
\begin{subfigure}{\columnwidth}
  \centering
  \includegraphics[width=\columnwidth]{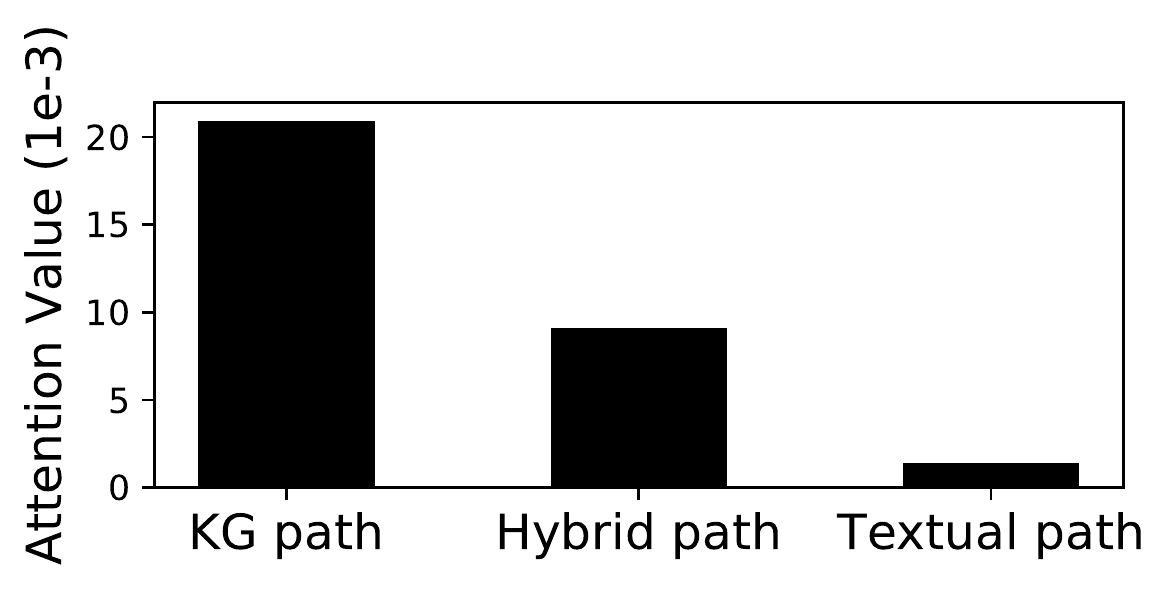}
  \caption{Path type and attention}
  \label{fig:attpathtp}
\end{subfigure}
\begin{subfigure}{\columnwidth}
  \centering
  \includegraphics[width=\columnwidth]{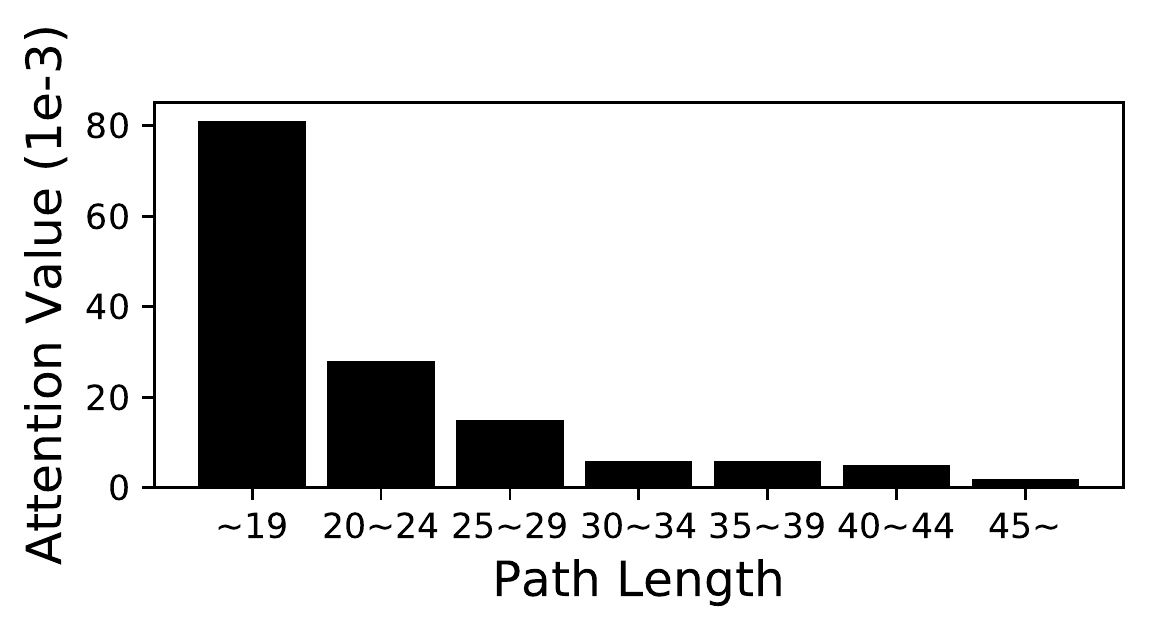}
  \caption{Path complexity and attention}
  \label{fig:attlen}
\end{subfigure}
\caption{(\textbf{Left}) Average attention weights across different path types. (\textbf{Right}) Average attention weights across different length range of paths. Here, we use path length as an indicator of path complexity, which denotes the number of words in a path.}
\label{fig:test}
\end{figure*}

While extending the base model into our UG setting, we observe that the base model tends to allocate more attention to KG or linguistically simple paths as compared to Textual paths (i.e., the path comes form Text), Hybrid paths (i.e., the path comes from both Text and KG), or linguistically complex ones, as shown in Figure~\ref{fig:attpathtp} and Figure~\ref{fig:attlen}. We consider that this would be because paths including Textual relations (i.e., Textual and Hybrid paths) or complex paths are comparatively noisier than KG or simple paths, but which does not necessarily mean the former is not useful. For instance, in Figure~\ref{fig:ugexam}, the complex Hybrid path $p_2$ is useful for predicting (\textbf{Colesevelam\_HCl}, \emph{may\_treat}, \textbf{Type\_2\_Diabetes}), because $p_2$ implies a plausible line of reasoning `` {\it \textbf{Colesevelam\_HCl} $\underrightarrow{alternative\_to}$ Colestipol $\underrightarrow{may\_treat}$ hyperglyceridemia $\underrightarrow{strong\_link\_to}$ \textbf{Type\_2\_Diabetes}}''. However, due to the attention bias mentioned above, the base model allocates low attention ($a'_2 \approx 8.0 \times 10^{-36}$) on the informative path, and thus fails to learn from such complex but useful evidences.

To reduce the negative effect of the attention biases and make full use of the UG path, we propose the following two training (or debiasing) strategies: Path Type Adaptive Pretraining (\S\ref{sec:pre}) and Complexity Ranking Guided Attention (\S\ref{sec:rank}).

\subsection{Path Type Adaptive Pretraining}\label{sec:pre} 
As shown in Figure~\ref{fig:attpathtp}, the base model tends to bias toward KG paths. This indicates that the base model mainly relies on KG paths so that it is incapable of capturing informative features from Textual and Hybrid paths.
This bias will decrease the flexibility and adaptability of the base model to different types of paths.

\begin{figure*}[t]
\centering
\includegraphics[width=13.5cm]{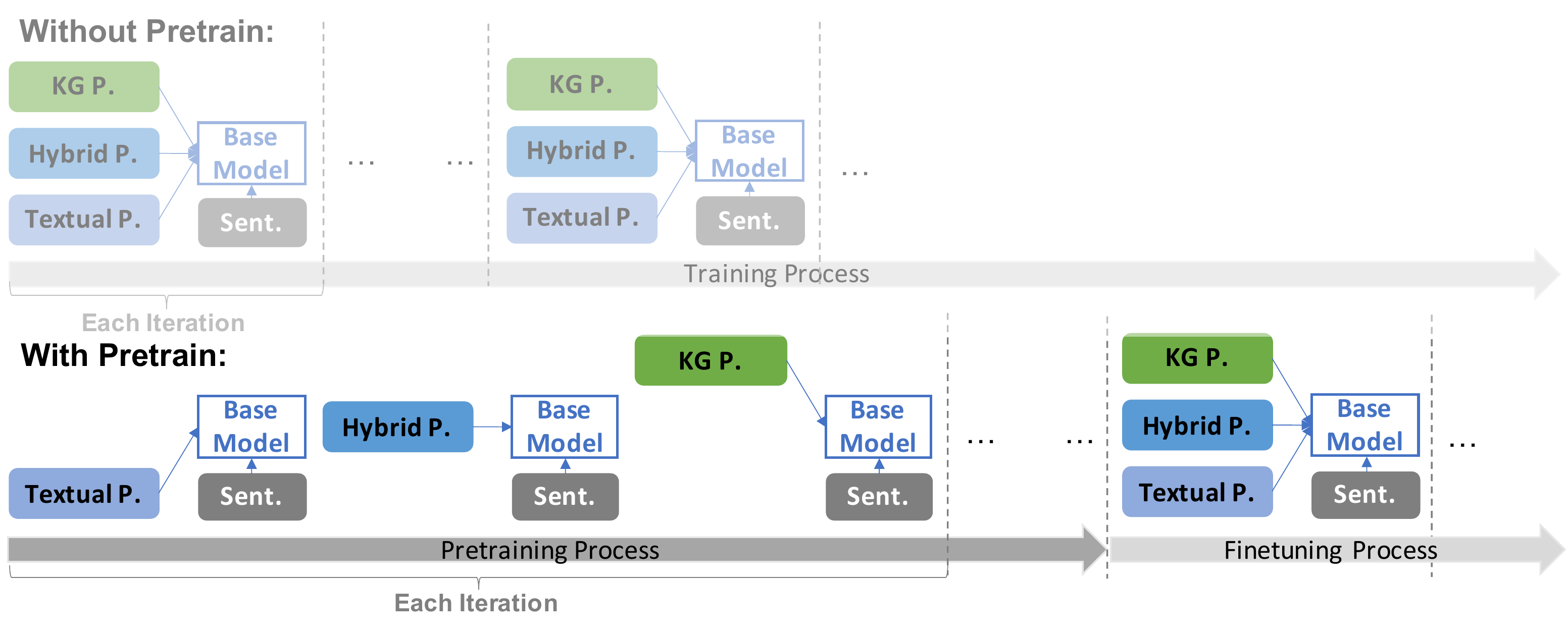}
\caption{Path Type Adaptive Pretraining strategy, where ``Textual/Hybrid/KG P.'' represent Textual, Hybrid, and KG paths respectively, and ``Sent.'' represents the sentence evidences. In this strategy (\textbf{Pretrain}), instead of using all types of paths to train the base model in all iterations, we sequentially train the model with Textual, Hybrid and KG paths, and then finetune it with all types of paths.}
\label{fig:pre_stra}
\end{figure*}

To address this issue, we propose a debiasing strategy called Path Type Adaptive Pretraining.
In this strategy, we pretrain the base model sequentially using Textual, Hybrid, and KG Paths as path evidences, and then finetune it with all types of paths as illustrated in Figure~\ref{fig:pre_stra}. We hypothesize that this strategy can prevent the reliance on a single type of UG path and improve the capacity of extracting features from the entire UG paths, and thereby increase the performance.

\subsection{Complexity Ranking Guided Attention}\label{sec:rank} 
Similar to the bias towards KG paths, the base model also focuses on linguistically simple paths, as shown in Figure~\ref{fig:attlen}, even though complex ones are informative (e.g. $p_2$ in Figure~\ref{fig:ugexam}). We hypothesize that 
restricting the attention span to the complex (simple) paths can force the model to pay attention to the complex (simple) paths, thereby effectively utilize them. Under this hypothesis, we propose a Complexity Ranking Guided Attention mechanism, as illustrated in Figure~\ref{fig:proposed_model}. 

Specifically, given a bag of paths $P_r=\{p_1,...,p_m\}$, we rank them according to their complexity scores ($\kappa $), which are calculated via $\kappa =\lambda_1\tau_1 + \lambda_2\tau_2 + ...$, where $\tau$ denotes the feature for capturing linguistic complexity (e.g., path length) and $\lambda$ is a corresponding weight, which is a hyperparameter. Sentence length (i.e., the number of tokens in a sentence) and lexical richness (i.e., the number of token types) are commonly used features for evaluating sentence complexity~\cite{brunato2018sentence}. Therefore, this work adopts them to calculate the complexity for a given path. 

Then, we group top $j$ most and least complex paths into a set of complex and simple paths respectively, where $j$ is a hyperparameter\footnote{In our experiments, we set $j$ as $30$ for NYT10 dataset and $50$ for Biomedical dataset.}. The set level representation is calculated via the Equation~\ref{eq:pathcomp}. 
\begin{equation}\label{eq:pathcomp}
\begin{split}
\mathbf{p}_{\mathrm{complex}\,or\,\mathrm{simple}}=\sum_{i\in top\_j \, or\, i\in last\_j}a'_i\mathbf{p}_i
\end{split}
\end{equation}
\begin{equation*}
\begin{split}
a'_i=\frac{\exp(\langle\mathbf{r}_{ht},\mathbf{x'}_i\rangle)}
{\sum_{k\in top\_j\,or\,k\in last\_j}\exp(\langle\mathbf{r}_{ht},\mathbf{x'}_k\rangle)},\\
\mathbf{x'}_i = \tanh(\mathbf{W}\mathbf{p}_i + \mathbf{b})
\end{split}
\end{equation*}
Finally, we concatenate the resulting representation $\mathbf{s}_{\mathrm{final}}$, $\mathbf{p}_{\mathrm{final}}$, $\mathbf{p}_{\mathrm{simple}}$ and $\mathbf{p}_{\mathrm{complex}}$ as the input to the relation classification layer. The conditional probability $P(e_1, r, e_2|S_r, P_r,\theta_S,\theta_P)$ is formulated via Equation~\ref{eq:pathall1}, where $\mathbf{o}=\mathbf{M}[\mathbf{s}_{\mathrm{all}}; \mathbf{p}_{\mathrm{all}};\mathbf{p}_{\mathrm{complex}};\mathbf{p}_{\mathrm{simple}}]+\mathbf{d}$. 
\begin{equation}\label{eq:pathall1}
P(e_1,r,e_2|S_r,P_r,\theta_S, \theta_P)=\frac{\exp([\mathbf{o}]_r)}
{\sum_{c=1}^{n_r}\exp([\mathbf{o}]_c)}
\end{equation}
%
%
%

%

\section{Experiments}

\subsection{Data}
\label{sec:data}
We evaluate our proposed framework on a biomedical dataset and NYT10 dataset~\cite{riedel2010modeling}. The statistics of both datasets is summarized in Table~\ref{tab:kb_sta}. We will detail both datasets as follows.

\textbf{Biomedical Dataset}. This datatset is created by linking biomedical KG with biomedical Text. We choose UMLS\footnote{\url{https://www.nlm.nih.gov/research/umls/}} and Medline corpus as the biomedical KG and Text respectively. UMLS is a frequently used biomedical knowledge base, while Medline corpus is a large collection of biomedical abstracts, both are developed and maintained by the U.S. National Library of Medicine\footnote{\url{https://www.nlm.nih.gov/}}. For identifying UMLS entity mentions in the Medline corpus, we use a state-of-the-art UMLS Named Entity Recognizer (NER), ScispaCy~\cite{neumann-etal-2019-scispacy}. The NER identifies UMLS concepts and annotates them by their corresponding UMLS Concept Unique Identifier (CUI) and entity types.

From the UMLS KG and the entity linked Medline corpus, we extract fact triplets (i.e., $(e_1, r, e_2)$) and corresponding sentence evidences containing $(e_1, e_2)$ under the restriction that: (1) each entity pair should be connected by a RO (RO stands for ``has Relationship Other than synonymous, narrower, or broader'') relationship; (2) each entity should belong to the following entity types: Protein, Gene, Disease or Syndrome, Enzyme, Chemical, Sign or Symptom and Pharmacologic Substance. Then we divide the collected triplets and sentence evidences into training and testing set according to the year when the source abstract of sentence evidence was published. The former is aligned to the years until 2008 and the latter to the years 2009 $\sim$ 2018, ensuring the testing set only contains the unobserved triplets. 

To simulate the noise in the real world, besides the ``related'' triplets, we also extract the ``unrelated'' triplets and sentence evidences based on a closed world assumption: pairs of entities not listed in a KG are regarded to have NA relation and sentences containing them are considered to be the NA sentence evidences. We divide the NA triplets and NA sentence evidences in the same way mentioned above. We use a subset of UMLS (see Appendix \S\ref{ap:sub})  and the Medline abstracts published until 2008 as the KG and Text respectively to create the UG for path retrieval. In addition, we use the same subset of UMLS triplets mentioned above to train the KG Encoder introduced in \S\ref{sec:evencoder}.

\textbf{NYT10}. The dataset is created by aligning Freebase relational facts with the New York Times Corpus. Sentence evidences from the year 2005 $\sim$ 2006 are used for training and the evidences from 2007 are used for testing. NYT10 dataset has been widely used by \cite{lin2016neural,ji2017distant,du2018multi,jat2018improving,du2018multi,han2018neural,han2018hierarchical,vashishth2018reside,ye2019distant,alt2019fine}. We use Freebase\footnote{From the entire Freebase, we only collect the triplets with the relations that are mentioned in NYT10 dataset for UG creation, ensuring not to overlap with testing set.} and ClueWeb12 with Freebase entity mention annotations~\cite{gabrilovich2013facc1} as the KG and Text to create the UG for path searching. In addition, following \cite{han2018neural}, we use FB60K for training the KG Encoder.
\begin{table}[t]
\centering
\scalebox{0.63}{
\begin{tabular}{c|c|c|c|c|c}
\hline
&\textbf{\#R} & \textbf{\#EP}& \textbf{\#Related EP} & \textbf{\#Sentence} &\textbf{\#UG Path} \\
\hline
\hline
Biomedical & 40 & \makecell{100,549 /\\ 21,081} & \makecell{10,936 /\\ 1,804} & \makecell{165,692 /\\ 28,912} & \makecell{12,854,696 /\\ 2,346,007}\\ \hline
NYT10 & 53 & \makecell{281,270 /\\ 96,678} & \makecell{18,252 /\\ 1,950} & \makecell{522,611 /\\  172,448} & \makecell{8,967,153 /\\  2,984,611} \\ \hline
\end{tabular}
}
\caption{Statistics of datasets in this work, where \textbf{R} and \textbf{EP} stand for the target Relation and Entity Pair, $\#_1$/$\#_2$ represent the number of training and testing data respectively.}
\label{tab:kb_sta}
\end{table}

\textbf{UG path search}. Given an entity pair ($e_1$, $e_2$), the UG path set $P_r$ is obtained by performing random walks over the UG from $e_1$ till $e_2$ with maximum step\footnote{We manually set the maximum step as 3.}.

\subsection{Settings}
\label{sec:settings}

We follow \cite{lin2016neural} and conduct the held-out evaluation, in which the model for DS-RE is evaluated by comparing the fact triplets identified from evidences (i.e., the bag of sentence evidences $S_r$ and the bag of UG path evidences $P_r$) with those in KG. Following the evaluation of previous works, we draw Precision-Recall curves and report the Area Under Curve (AUC) and Precision@N (P@N) metrics, which gives the percentage of correct triplets among top N ranked candidates.
The parameter settings of our experiments are detailed in Appendix \S\ref{ap:para}.

\begin{figure*}[t]
\centering
\begin{minipage}{.5\textwidth}
  \centering
  \includegraphics[width=\columnwidth]{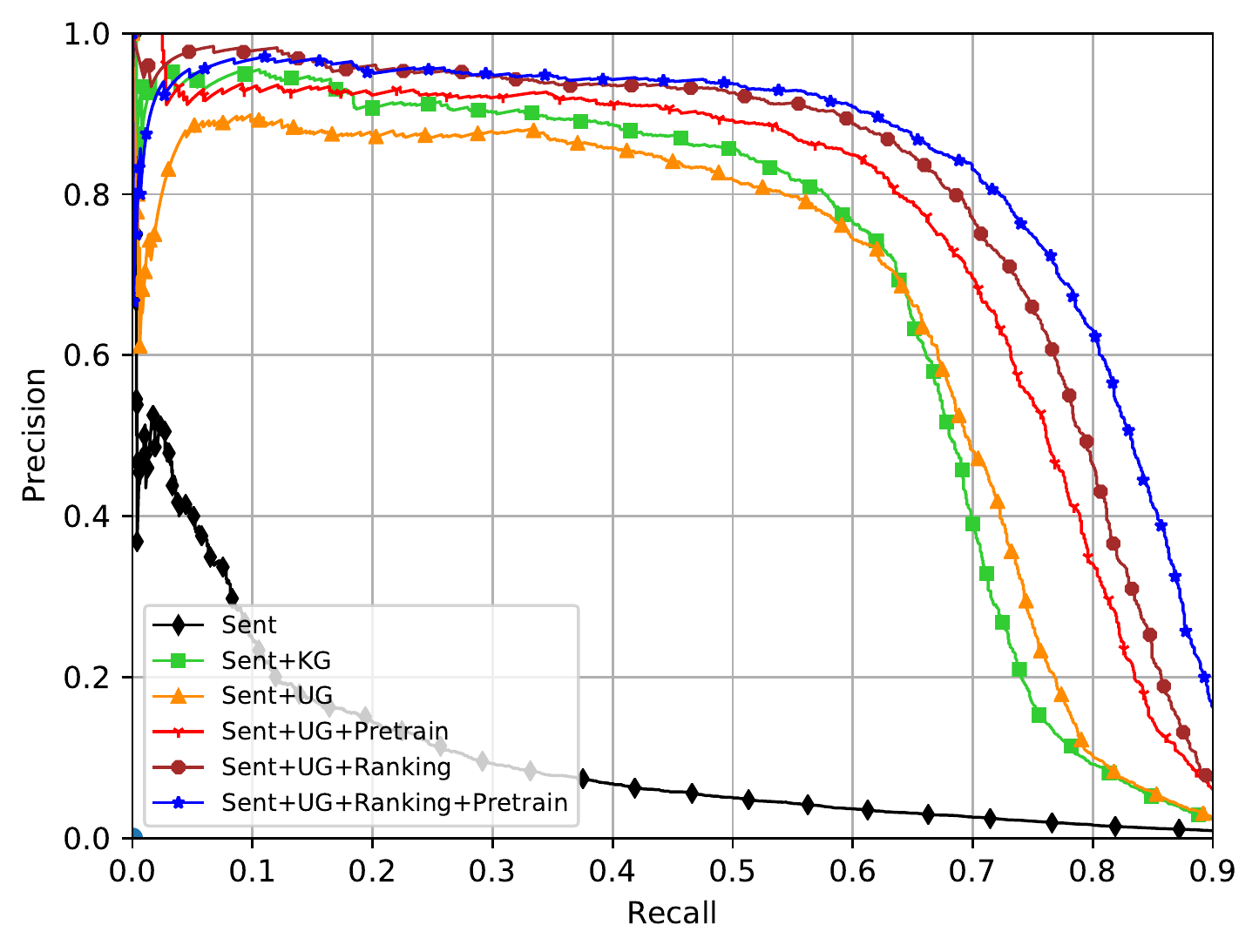}
  \captionof{figure}{PR curves on Biomedical dataset.}
  \label{fig:prc_bio}
\end{minipage}%
\begin{minipage}{.5\textwidth}
  \centering
  \includegraphics[width=\columnwidth]{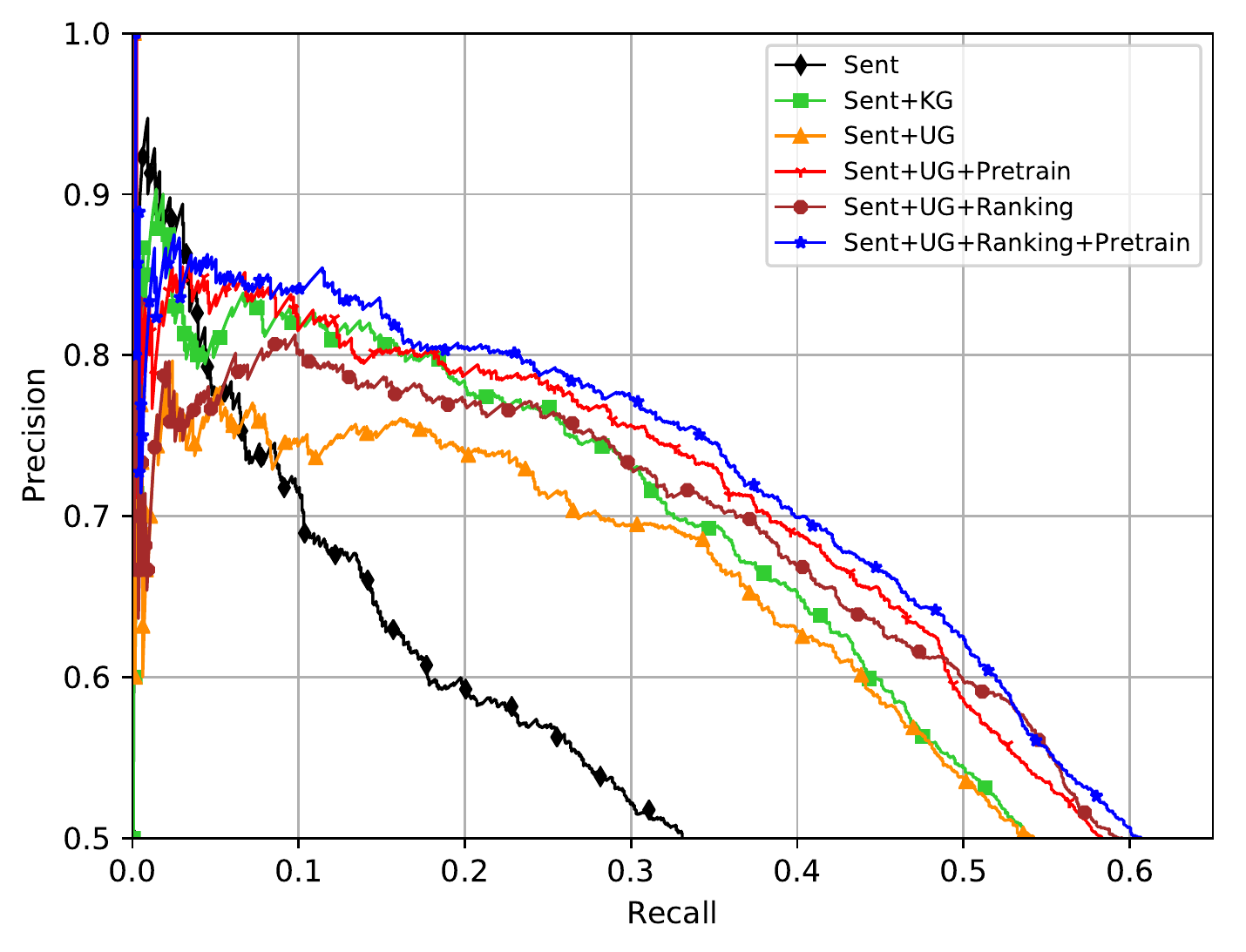}
  \captionof{figure}{PR curves on NYT10 dataset.}
  \label{fig:prc_nyt}
\end{minipage}
\end{figure*}

\begin{table*}[t]
\centering
\scalebox{0.78}{
\begin{tabular}{lccccccccccccccc}
\hline
&\multicolumn{5}{c}{\textbf{Biomedical dataset}} &\multicolumn{7}{c}{\textbf{NYT10 dataset}}\\\cmidrule(lr){2-6}\cmidrule(lr){7-13}
\textbf{Model} & AUC & P@0.5k & P@1k & P@1.5k & P@2k & AUC & P@0.1k & P@0.2k & P@0.3k & P@0.5k & P@1k & P@2k\\\cmidrule(lr){1-6}\cmidrule(lr){7-13}\cmidrule(lr){1-6} \cmidrule(lr){7-13}
Sent & 9.6 & 30.0 & 20.8 & 18.7 & 16.3 & 36.6 & 81.0 & 73.5 & 68.3 & 62.0 & 53.8 & 40.2 \\\cmidrule(lr){1-6}\cmidrule(lr){7-13}
Sent+KG & 62.6 & 91.4 & 86.1 & 74.2 & 58.5 & 50.2 & 80.0 & 82.0 & 81.3 & 77.2 & 67.9 & 50.3\\\cmidrule(lr){1-6}\cmidrule(lr){7-13}
Sent+UG & 61.0 & 87.6 & 83.4 & 73.8 & 58.5 & 48.4 & 74.0 & 76.0 & 74.7 & 74.0 & 66.7 & 50.3\\\cmidrule(lr){1-6}\cmidrule(lr){7-13}
\makecell[l]{Sent+UG\\+Pretrain} & 70.1 & 95.4 & 89.7 & 76.3 & 60.4 & 52.7 & 83.0 & 82.0 & 80.3 & 78.6 & 70.4 & 52.6 \\\cmidrule(lr){1-6}\cmidrule(lr){7-13}
\makecell[l]{Sent+UG\\+Ranking} & 74.2 & 95.2 & 92.2 & 81.1 & 62.2 & 52.1 & 77.0 & 80.0 & 79.3 & 77.4 & 70.3 & 54.4 \\\cmidrule(lr){1-6}\cmidrule(lr){7-13}
\makecell[l]{Sent+UG\\+Ranking\\+Pretrain} & \textbf{77.5} & \textbf{95.4} & \textbf{93.1} & \textbf{83.9} & \textbf{64.4} & \textbf{55.0} & \textbf{86.0} & \textbf{84.0} & \textbf{83.3} & \textbf{80.4} & \textbf{71.9} & \textbf{54.5} \\\hline
\end{tabular}
}
\caption{P@N and AUC on Biomedical and NYT10 dataset (k=1000).}
\label{tab:patn_bio_nyt}
\end{table*}
\begin{figure}[t]
\centering
\includegraphics[width=\columnwidth]{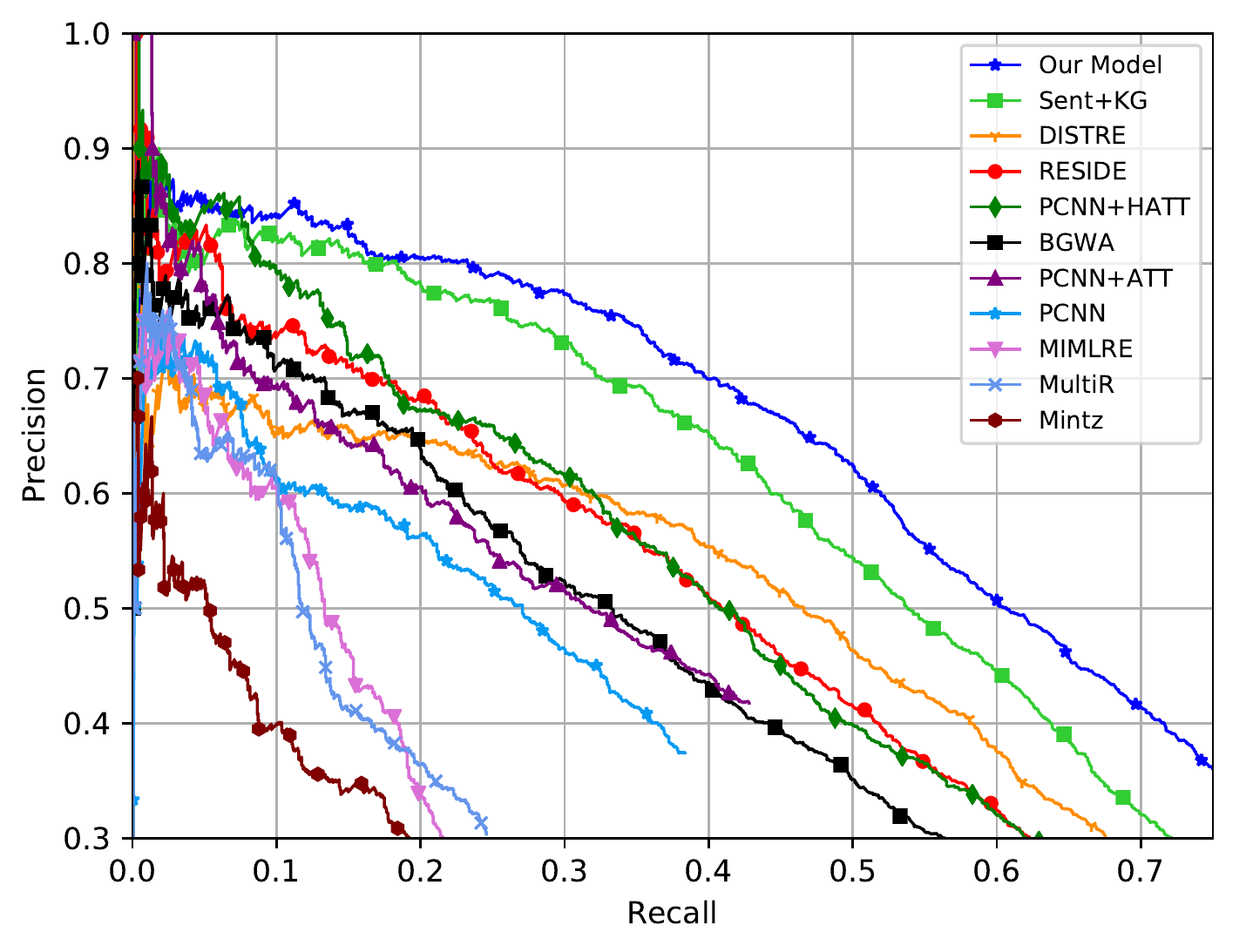}
\caption{PR curves of previous state-of-the-art methods and our proposed model on NYT10 dataset.}
\label{fig:prc_stnyt}
\end{figure}
\begin{table*}[t]
\centering
\scalebox{0.8}{
\begin{tabular}{c c c c c c c c}
\hline
\textbf{System} & AUC & P@0.1k & P@0.2k & P@0.3k & P@0.5k & P@1k & P@2k\\\hline\hline
Mintz\dag & 10.7 & 52.3 & 50.2 & 45.0 & 39.7 & 33.6 & 23.4 \\ \hline
PCNN+ATT\dag & 34.1 & 73.0 & 68.0 & 67.3 & 63.6 & 53.3 & 40.0 \\ \hline
RESIDE\dag & 41.5 & 81.8 & 75.4 & 74.3 & 69.7 & 59.3 & 45.0 \\ \hline
PCNN+HATT\ddag & 42.0 & 81.0 & 79.5 & 75.7 & 68.0 & 58.6 & 42.1 \\ \hline
DISTRE\dag & 42.2 & 68.0 & 67.0 & 65.3 & 65.0 & 60.2 & 47.9 \\ \hline
Sent+KG & 50.2 & 80.0 & 82.0 & 81.3 & 77.2 & 67.9 & 50.3 \\ \hline
Our Model & \textbf{55.0} & \textbf{86.0} & \textbf{84.0} & \textbf{83.3} & \textbf{80.4} & \textbf{71.9} & \textbf{54.5} \\ \hline
\end{tabular}
}
\caption{P@N and AUC from previous state-of-the-art DS-RE models and our proposed model on NYT10 dataset, where \dag represents that these results are quoted from \cite{alt2019fine} and \ddag indicates the results using the pretrained model from \cite{han2018hierarchical}.}
\label{tab:patn_stnyt}
\end{table*}

To demonstrate the effectiveness of our framework, we choose the model proposed by \citet{dai2019incorporating} as the baseline model, because this is the closest model in terms of incorporating multiple paths for DS-RE. Henceforth, ``Sent+KG'' is the baseline model, which uses both sentence evidences and KG paths. ``Sent+UG'' represents the base model in \S\ref{sec:evencoder} which takes UG paths instead of KG paths as path evidences. ``Sent+UG+Pretrain'' and ``Sent+UG+Ranking'' denote the base model trained with Path Type Adaptive Pretraining strategy and the base model with Complexity Ranking Guided Attention mechanism, respectively.
``Sent+UG+Ranking+Pretrain'' means the base model trained with both strategies.

\subsection{Results and Discussion}
\label{sec:results}

\textbf{Precision-Recall Curves}. The Precision-Recall (PR) curves of each model on the biomedical and NYT10 datasets are shown in Figure~\ref{fig:prc_bio} and Figure~\ref{fig:prc_nyt}, respectively.
The results show that: (1) ``Sent+UG'' does not have obvious advantages than ``Sent+KG'', illustrating that due to the biases discussed in \S\ref{sec:attbia}, simply applying UG paths on the base model has limited effect on improving the performance of DS-RE. (2) ``Sent+UG+Pretrain'' and ``Sent+UG+Ranking'' achieve better overall performance than ``Sent+KG'' on both datasets, especially when the recall is greater than $0.3$, demonstrating that UG has the potential to enhance the performance and the two proposed debiasing strategies are effective for exploiting the potential of UG for DS-RE. (3) ``Sent+UG+Ranking+Pretrain'' achieves the highest precision over the (almost) entire recall range on both datasets, proving that the two proposed strategies have a mutual complementary relationship on exploiting UG for DS-RE. This is understandable because the two proposed strategies deal with different types of biases, in addition, ``Pretrain'' helps the base model adapt to UG paths by effectively tuning its weights, while ``Ranking'' enhances the base model by adjusting its attention mechanism. (4) The consistent improvement on two datasets from different domains further proves the validity of our proposed methods. 

\textbf{AUC and P@N Evaluation}. Table~\ref{tab:patn_bio_nyt} further presents the results in terms of AUC and P@N. From them, we have similar observation to the PR curves.
We also observe that the effectiveness of UG paths is more pronounced on Biomedical dataset than on NYT10 dataset. We speculate that compared to the generic NYT10 dataset, further Background Knowledge (BK) is needed to identify relations from Biomedical dataset, and UG paths could be utilized as the BK to facilitate the scientific DS-RE.

\textbf{Comparison with State-of-the-art Baselines on NYT10}. To demonstrate the effectiveness of our proposed model, we also compare it against the following baselines on NYT10 dataset: Mintz~\cite{mintz2009distant}, MultiR~\cite{hoffmann2011knowledge}, MIMLRE~\cite{surdeanu2012multi}, PCNN~\cite{zeng2015distant}, PCNN+ATT~\cite{lin2016neural}, BGWA~\cite{jat2018improving}, PCNN+HATT~\cite{han2018hierarchical}, RESIDE~\cite{vashishth2018reside}, DISTRE~\cite{alt2019fine} and Sent+KG~\cite{dai2019incorporating}.
The results shown in Figure~\ref{fig:prc_stnyt} and Table~\ref{tab:patn_stnyt} indicate that: (1) our selected base model, ``Sent+KG'', is a strong baseline because it significantly outperforms other state-of-the-art models; and (2) our model can effectively take advantage of the rich UG paths for DS-RE because it beats the strong baseline and achieves a new state-of-the-art result on the commonly used DS-RE dataset.

\begin{table}[ht]
\centering
\resizebox{\columnwidth}{3.0cm}{
\begin{tabular}{cccl}
\hline
Base & Prop. & \textbf{Biomedical Triplet} \\ \hline
\xmark & \cmark & ( \colorbox{green!30}{Beta-2...Gene}, gene\_associated\_with\_disease, \colorbox{red!30}{Asthma})\\ \hline
\multicolumn{2}{c}{} & \textbf{Multi-hop Path} \\ \hline
Low & \textbf{High} & \makecell[l]{$hop_1$: \it ``The human \colorbox{green!30}{Beta-2...Gene} is responsible for \\ \it the binding of endogenous \colorbox{black!10}{Catecholamine} and their ...'' \\ $hop_2$: \it ``\colorbox{black!10}{Catecholamine} chemical structure of \colorbox{orange!30}{Epinephrine}'' \\ $hop_3$: \it `` \colorbox{orange!30}{Epinephrine} may treat \colorbox{red!30}{Asthma}''.} \\  \hline\hline
Base & Prop. & \textbf{NYT10 Triplet} \\ \hline
\xmark & \cmark & ( \colorbox{green!30}{San\_Francisco}, /location/contains, \colorbox{red!30}{Noe\_Valley})\\ \hline
\multicolumn{2}{c}{} & \textbf{Multi-hop Path} \\ \hline
Low & \textbf{High} & \makecell[l]{$hop_1$: \it ``\colorbox{green!30}{San\_Francisco} /location/contains \colorbox{black!10}{Fort\_Point}'' \\ $hop_2$: \it ``Surf spots and surfing regions include Northern CA, \\ \it the \colorbox{orange!30}{Bay\_Area}, San Francisco, Ocean Beach and \colorbox{black!10}{Fort Point}'' \\ $hop_3$: \it ``\colorbox{orange!30}{Bay\_Area} /location/contains \colorbox{red!30}{Noe\_Valley}''} \\  \hline
\end{tabular}
}
\caption{\label{tab:casestudy} Some examples of attention distribution over paths from ``Sent+UG'' (Base) and ``Sent+UG+Ranking+Pretrain'' (Prop.), where \cmark (or \xmark) represents the correct (or incorrect) prediction of the target relation.}
\end{table}

\textbf{Case Study}. Table~\ref{tab:casestudy} shows the UG path examples that are scored with highest (``High'') or lowest (or lower than $1.0\times10^{-3}$) (``Low'') attention by the base model and our proposed framework. The paths in the table generally mean ``\colorbox{green!30}{\it Beta-2... Gene} $\underrightarrow{is\_responsible\_for}$ \colorbox{black!10}{\it Catecholamine} $\underrightarrow{is\_the\_chemical\_class\_of}$ \colorbox{orange!30}{\it Epinephrine} $\underrightarrow{may\_treat}$ \colorbox{red!30}{\it Asthma}'' and ``\colorbox{green!30}{\it San\_Francisco} $\underrightarrow{contains}$ \colorbox{black!10}{\it Fort\_Point} $\underrightarrow{equal\_status}$ \colorbox{orange!30}{\it Bay\_Area} $\underrightarrow{contains}$ \colorbox{red!30}{\it Noe\_Valley}'', and thus can be seen as the useful path evidences for identifying \emph{gene\_associated\_with\_disease} and \emph{/location/contains} relation respectively. These examples indicate that our proposed training strategies could help the base model attend such informative UG paths so that it can correctly identify the target relation.

\section{Conclusion and Future Work}
We have introduced UG paths as extra evidences for the task of DS-RE from text. In order to fully take advantage of the rich UG paths, we have proposed two training (or debiasing) strategies: Path Type Adaptive Pretraining and Complexity Ranking Guided Attention mechanism. We have conducted experiments on both biomedical and NYT10 datasets. The results show that the two proposed methods are effective for exploiting the potential of UG paths for improving the performance of DS-RE. 

In the future, we plan to carry out the following steps: (1) we further investigate how the proposed training methods influence the performance via manual analysis so as to better the efficiency; and (2) instead of random walk, we may collect UG paths by adopting more sophisticated mechanisms such as training a path searching agent via reinforcement learning to prevent redundant and noisy paths.  

\section*{Acknowledgement}
This work was supported by JST CREST Grant Number JPMJCR1513, Japan and KAKENHI Grant Number 16H06614. We would like to thank Benjamin Heinzerling, Shota Sasaki and other collaborators for their useful comments and suggestions.

\bibliography{eacl_2021}
\bibliographystyle{acl_natbib}

\newpage
\appendix
\section{Appendix}
\subsection{CNN-Max}\label{ap:cnnmax}
Convolutional Neural Network with Max pooling layer (CNN-Max) is adopted to derive the sentence representation $\mathbf{s}$ and path representation $\mathbf{p}$. Specifically, vector representation $\mathbf{v}_t$ for each word $w_t$ is calculated via Equation~\ref{eq:cnnemb1}, where $\mathbf{W}_{emb}^w$ is a word embedding projection matrix~\cite{mikolov2013distributed}, $\mathbf{W}_{emb}^{wp}$ is a word position embedding projection matrix~\cite{zeng2014relation}, $\mathbf{x}_t^w$ is a one-hot word representation and $\mathbf{x}_t^{wp}$ is a one-hot word position representation, which indicates the relative distance between the current word and the target entity pair. 
\begin{equation}\label{eq:cnnemb1}
\mathbf{v}_t=[\mathbf{v}_t^w; \mathbf{v}_t^{wp1}; \mathbf{v}_t^{wp2}],
\end{equation}
\begin{equation*}
\begin{split}
\mathbf{v}_t^w=\mathbf{W}_{emb}^w\mathbf{x}_t^w, \\
\mathbf{v}_t^{wp1}=\mathbf{W}_{emb}^{wp}\mathbf{x}_t^{wp1}, \\
\mathbf{v}_t^{wp2}=\mathbf{W}_{emb}^{wp}\mathbf{x}_t^{wp2}
\end{split}
\end{equation*}
The sentence representation $\mathbf{s}$ and path representation $\mathbf{p}$ are formulated via the Equation~\ref{eq:cnnmax}, where $\mathbf{W}^{sent}$ (or $\mathbf{W}^{path}$) is the convolution kernal, $\mathbf{b}^{sent}$ (or $\mathbf{b}^{path}$) is the corresponding bias vector, $\mathbf{v}^{sent}_t$ (or $\mathbf{v}^{path}_t$) is the vector for each word $w_t$ in a sentence (or path), $[vec]_i$ is the $i$-th value of $vec$, $\nu$ is the dimensionality of $\mathbf{s}$ and $\mathbf{p}$, and $k$ is the convolutional window size.
\begin{equation}\label{eq:cnnmax}
\begin{split}
[\mathbf{s}]_i = \max\limits_{t}\{[\mathbf{h}^{sent}_{t}]_{i}\},\ \forall i=1, ..., \nu \\
[\mathbf{p}]_i = \max\limits_{t}\{[\mathbf{h}^{path}_{t}]_{i}\},\ \forall i=1, ..., \nu \\
\end{split}
\end{equation}
\begin{equation*}
\begin{split}
\mathbf{h}^{sent}_t=\tanh(\mathbf{W}^{sent}\mathbf{z}^{sent}_t + \mathbf{b}^{sent}),\\
\mathbf{h}^{path}_t=\tanh(\mathbf{W}^{path}\mathbf{z}^{path}_t + \mathbf{b}^{path}),\\
\mathbf{z}^{sent}_t=[\mathbf{v}^{sent}_{t-(k-1)/2}; ... ; \mathbf{v}^{sent}_{t+(k-1)/2}],\\
\mathbf{z}^{path}_t=[\mathbf{v}^{path}_{t-(k-1)/2}; ... ; \mathbf{v}^{path}_{t+(k-1)/2}]
\end{split}
\end{equation*}

\subsection{Parameter Settings}\label{ap:para}
All of the hyperparameters used in our experiments are listed in Table~\ref{tab:hypara}. Most of them follow the hyperparameter setting in \cite{dai2019incorporating} and \cite{han2018neural}. We use a Word2Vec model\footnote{Gensim word2vec implementation: \url{https://radimrehurek.com/gensim/models/word2vec.html}} to train the word embeddings on the UMLS entity linked corpus for the biomedical dataset, and adopt the word embeddings released by \cite{lin2016neural} for NYT10 dataset. We apply Stochastic Gradient Descent (SGD) to optimize the proposed DS-RE model.
\begin{table}[t]
\centering
\scalebox{0.8}{
\begin{tabular}{|l|c|c|}
\hline
\textbf{Hyperparameter} & \textbf{Biomedical} & \textbf{NYT10} \\\hline
word embedding dimension & 50 & 50 \\ \hline
KG embedding dimension & 50 & 50 \\ \hline
position embedding dimension & 5 & 5 \\ \hline
CNN window size & 3 & 3 \\ \hline
CNN filter number & 100 & 230 \\ \hline
dropout rate & 0.5 & 0.5 \\ \hline
\makecell[l]{learning rate \\(for sentences and paths)} & 0.02 & 0.05 \\ \hline
learning rate (for KG) & 0.05 & 0.001 \\ \hline
batch size & 50 & 160 \\ \hline
\end{tabular}
}
\caption{Hyperparameters used in our experiments.}
\label{tab:hypara}
\end{table}

\subsection{Subset of UMLS}\label{ap:sub}
Besides the 7 entity types mentioned above, we also use other 22 entity types, as listed in Table~\ref{tab:umls_tp}, to collect the UMLS triplets that are connected by RO relationship, ensuring all testing triplets are removed. The main reasons to manually restrict the entity type is because (1) we observe that most of the Medline abstracts discuss the relationship among these entity types; (2) these concrete entities could prevent semantic drift while searching UG paths.

\begin{table}[t]
\centering
\scalebox{1.0}{
\begin{tabular}{c}
\hline
\textbf{Selected Entity Types} \\\hline
\makecell[l]{Antibiotic, Biologically Active Substance, \\ Bacterium, Organ, Cell Component,\\ Cell Function, Cell, Clinical Drug, Ion, \\ Eukaryote, Food, Genetic Function \\ Hazardous or Poisonous Substance, \\ Hormone, Immunologic Factor,\\ Inorganic Chemical, Organic Chemical, \\ Pathologic Function, Receptor, \\ Steroid, Virus and Vitamin.} \\ \hline
\end{tabular}
}
\caption{List of selected UMLS entity types.}
\label{tab:umls_tp}
\end{table}

\end{document}